\documentclass[letterpaper, 10 pt, conference]{ieeeconf}  
\usepackage{graphicx}
\usepackage{amsmath}

\IEEEoverridecommandlockouts                              

\title{\LARGE \bf
Steerable rolling of a 1-DoF robot using an internal pendulum
}

\author{{Christopher Y. Xu$^1$, Jack Yan$^2$, Kathleen Lum$^1$, and Justin K. Yim$^2$}
\thanks{$^{1,2}$The authors are with the $^1$Department of Electrical and Computer Engineering and the $^2$Department of Mechanical Science and Engineering at the University of Illinois at Urbana-Champaign, USA. 
\texttt{jkyim@illinois.edu}}
}

\begin{document}

\maketitle
\thispagestyle{empty}
\pagestyle{empty}

\begin{abstract}

We present ROCK (Rolling One-motor Controlled rocK), a 1 degree-of-freedom robot consisting of a round shell and an internal pendulum. An uneven shell surface enables steering by using only the movement of the pendulum, allowing for mechanically simple designs that may be feasible to scale to large quantities or small sizes. We train a control policy using reinforcement learning in simulation and deploy it onto the robot to complete a rectangular trajectory. 
\vspace{+0.5cm}
\end{abstract}

\section{INTRODUCTION}

\subsection{Motivation}
Existing spherical robot designs require two to four actuators for steering and jumping capabilities \cite{ARMOUR2006195}, increasing cost, power use, maintenance, and size. Reducing the number of actuators can alleviate these challenges, making it easier to scale to greater numbers and smaller sizes.

This work investigates ROCK, a robot with an internal pendulum controlled by a single motor capable of rolling, steering, and jumping. ROCK achieves locomotion by shifting its center of mass as the pendulum rotates and steers by modulating motor acceleration.

\subsection{Prior work}
 Our approach integrates an offset pendulum, custom shell design, and reinforcement learning to support controlled steering with a single motor.
 
 Wheels, levers, and rotating and linear pendulums have been explored as mechanisms to achieve locomotion in spherical robots with low actuator count. These solutions leverage changes in a system's moment of inertia and conservation of momentum to move \cite{ARMOUR2006195} and guide our investigation of simple mechanisms for ground locomotion. 
 
Compressible or non-spherical shells have also demonstrated improved performance on rough terrain offering better traction and reducing unwanted rolling \cite{9762154}. 
 
The interaction of the uneven shell surface with the ground results in complex dynamics. To enable reliable steering, reinforcement learning (RL) is used to simulate various terrains and reward systems, enabling it to learn dynamic movement patterns suitable for various terrains \cite{7989079}. 
\vspace{+0.5cm}

\section{METHODS}

\begin{figure}
    \centering
    \includegraphics[width=\columnwidth]{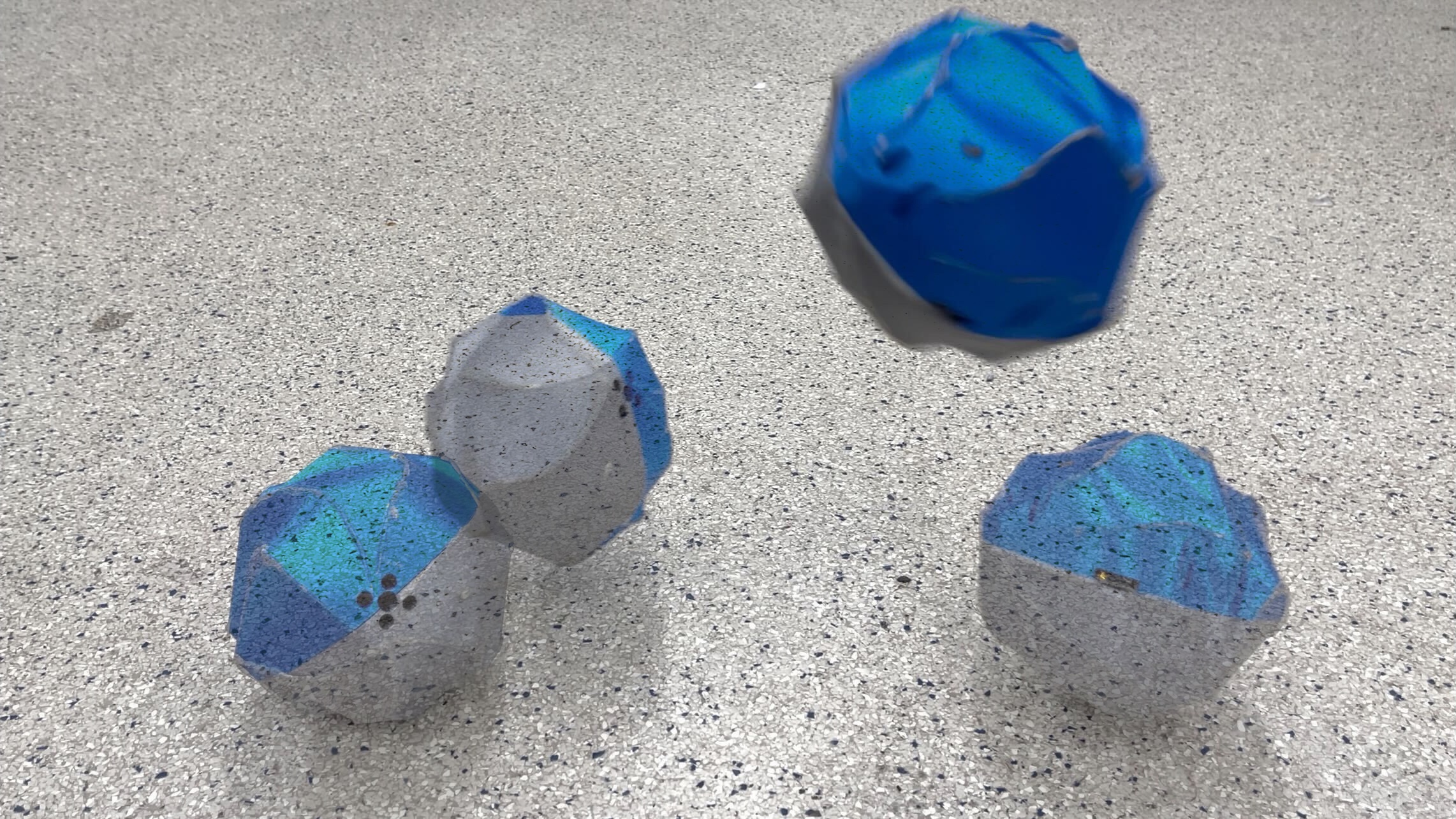}
    \caption{Overlaid snapshots of rock turning and jumping.}
    \label{fig:intro}
    \vspace{-0.2cm}
\end{figure}
\subsection{Hardware}

\subsubsection{Outer shell design}

We avoid a perfectly spherical shell since its extreme symmetry leaves few affordances for steering left and right. The shell design is divided into two identical hemispheres that are joined with a 180 degree rotational offset, with each hemisphere having one bulged side and one dented side, so that the robot tilts left and right as it rolls on the ground. As contact with the ground changes, actuating the motor results in varying ground reaction forces that create a torque to steer the robot. 

\begin{figure}[b]
    \centering
    \includegraphics[width=\columnwidth]{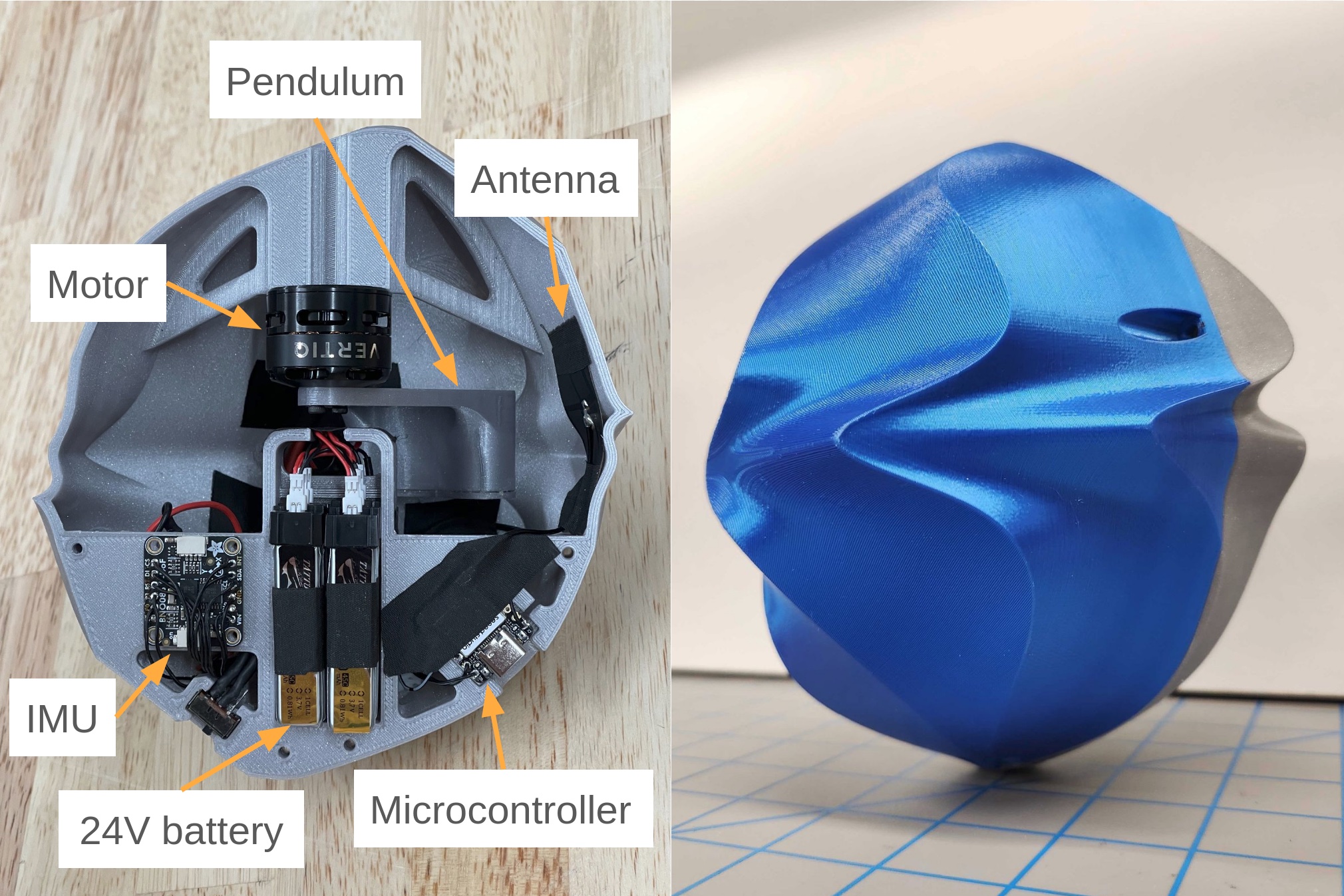}
    \caption{\textit{Left}: Internals of ROCK, showing the electronics and pendulum. \textit{Right}: Uneven exterior plastic shell of ROCK.}
    \label{fig:enter-label}
\end{figure}

\subsubsection{Electronics}
We use an ESP32-S3 microcontroller for onboard processing and Wi-Fi communication, a BNO085 IMU for orientation sensing, six Tattu 220mAh LiPo cells to power the electronics and motor, and a Vertiq 23-06 220Kv motor to rotate the pendulum for locomotion. A 6mm coreless brushed DC motor with a 30mm diameter propeller provides forced air cooling to the Vertiq motor but does not contribute to locomotion. 

\subsection{Control}
The robot can be teleoperated by commanding a desired rolling direction. We first implemented a baseline controller that rolls by offsetting the robot's center of mass in the direction by manual joystick input, then trained a learning-based controller to use dynamic pendulum movements for higher performance.

\subsubsection{Projection controller}

\begin{figure}[h]
    \vspace{-1em}
    \centering
    \includegraphics[width=\columnwidth]{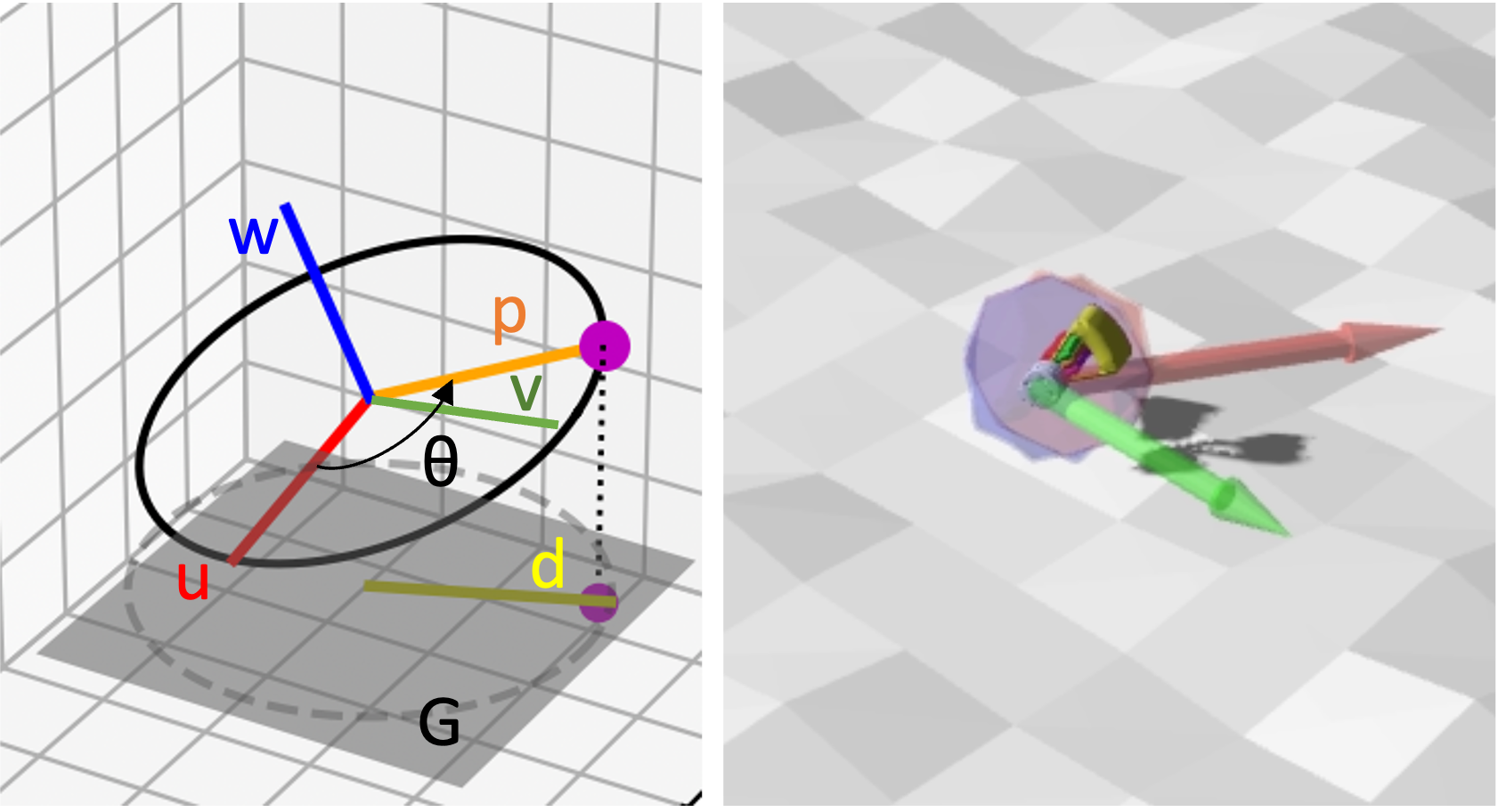}
    \caption{\textit{Left:} The projection controller computes pendulum angle $\theta$ such that its projection onto the horizontal plane is in commanded direction $\textbf{d}$. \textit{Right:} A policy is trained in simulation to track a randomly sampled commanded direction on rough terrain.}
    \label{fig:controller}
\end{figure}

We first develop a simple controller to place the pendulum at an angle that moves the center of mass in the desired rolling direction. The pendulum angle is chosen such that its projection onto the horizontal plane is in the direction of the commanded locomotion direction, satisfying
\begin{align} \label{eq:projection}
    proj_{G}(\textbf{p}) = \lambda\textbf{d}, \quad   \lambda>0
\end{align}
where $G$ denotes the ground plane, \textbf{p} is the pendulum center of mass displacement from the system centroid, and \textbf{d} is a unit vector specifying the commanded direction, visualized in Fig. \ref{fig:controller}. Vectors $\textbf{u, v, w}$ specify the motor frame, with $\textbf{w}$ pointing out of the motor. The pendulum angle $\theta$ with respect to $\textbf{u}$ is computed and can be tracked with a PD controller.

\vspace{-1em}
\begin{align}
   \theta = \begin{cases}
    \operatorname{atan2} \left(u_x d_y - u_y d_x, v_y d_x - v_x d_y \right) & \text{if } w_z \geq 0 \\
    \operatorname{atan2} \left(u_x d_y - u_y d_x, v_y d_x - v_x d_y \right) + \pi       & \text{otherwise}
    \end{cases}
\end{align}

\subsubsection{Learning-based controller}
The projection controller enables directional rolling by shifting its center of mass, but it does not consider the uneven surface or the reaction forces generated by accelerating the pendulum. To take advantage of the dynamics, we trained a reinforcement learning policy using proximal policy optimization (PPO) in the Genesis physics simulator \cite{genesis}. A MLP policy with fully connected hidden layers [512, 256, 128] is trained in Pytorch then converted to TensorFlow-Lite for the microcontroller. Each observation is scaled and clipped to have magnitude 1, allowing the model to be quantized using int8 and computed with SIMD optimizations for inference in under 8 ms. 

The observations are shown in Table \ref{tab:obs}. The past two observations also used as inputs, resulting in an input shape of 45. The output is a single value constrained within [-1, 1] that is mapped to a target motor velocity within $\pm$21 rad/s.

\begin{table}[h]
    \centering
    \begin{tabular}{|l|c|c|}
        \hline
        Observation & Shape & Scale\\
        \hline
        Last action & 1 & 1\\
        Command direction & 2 & 1\\
        Body orientation quaternion & 4 & 1\\
        Body angular velocity $\textbf{w}$ & 1 & 1/24 rad/s\\
        Body angular velocity $\textbf{u, v}$ & 2 & 1/12 rad/s\\
        Motor angular velocity & 1 & 1/37.5 rad/s\\
        sin, cos of motor angle & 2 & 1\\
        sin, cos of proj. angle $\theta$ &2 &1\\
        \hline
    \end{tabular}
    \caption{Observations for reinforcement learning policy}
    \label{tab:obs}
    \vspace{-2em}
\end{table}

\section{RESULTS}

\subsection{Trajectory following}

A human operator controls the robot to follow a rectangular 2 x 3 m trajectory by commanding the desired rolling direction through a wireless joystick. The robot is able to complete the trajectory by tracking the desired direction using the learning-based controller.

\begin{figure}[h]
    \centering
    \includegraphics[width=\columnwidth]{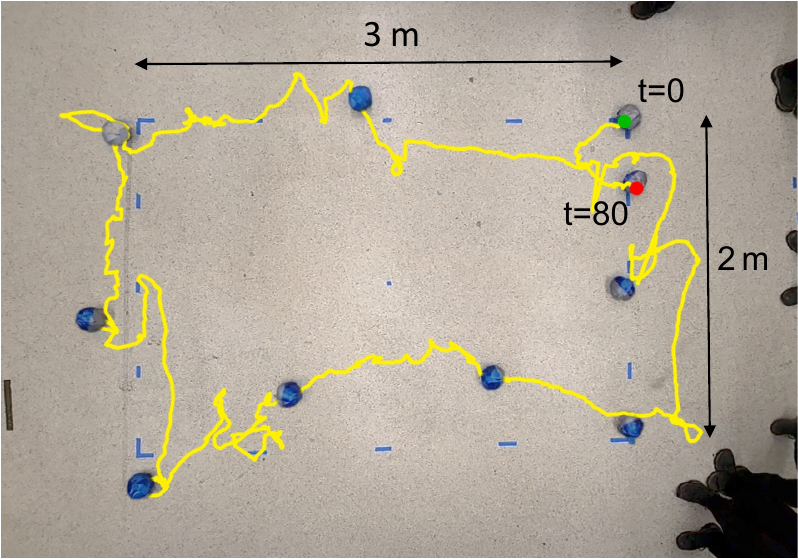}
    \caption{Overhead view: A user controls the robot to follow a rectangular trajectory marked by blue tape on the ground. The actual path of the robot is shown in yellow. The trajectory is completed in 80 seconds, resulting in an average speed of about 0.13 m/s.}
    \label{fig:path}
    \vspace{-1em}
\end{figure}

\subsection{Jumping}
Similar to  \cite{buzhardt2023pendulum}, ROCK can jump by swinging its pendulum up vertically, generating a reaction force on the shell that, in turn, pushes on the ground (Fig. \ref{fig:intro}).

\addtolength{\textheight}{-12cm}   






\bibliographystyle{IEEEtran}
\bibliography{references.bib}

\end{document}